\documentclass[letterpaper,journal]{IEEEtran}
\usepackage{amsmath,amsfonts}
\usepackage{algorithmic}
\usepackage{algorithm}
\usepackage{array}
\usepackage[caption=false,font=normalsize,labelfont=sf,textfont=sf]{subfig}
\usepackage{textcomp}
\usepackage{stfloats}
\usepackage{siunitx}
\usepackage{float}
\usepackage{url}
\usepackage{xurl}
\usepackage{verbatim}
\usepackage{graphicx}
\usepackage{cite}
\usepackage[dvipsnames]{xcolor}
\usepackage{booktabs}
\usepackage{multirow}   
\usepackage{tabularx}
\usepackage{hyperref}
\usepackage{gensymb}
\usepackage{todonotes}

\usepackage{xcolor}
\definecolor{darkgreen}{rgb}{0.0, 0.5, 0.0}

\newif\ifrevisions
\revisionsfalse
\newcommand{\revisionadd}[1]{%
  \ifrevisions
    \textcolor{darkgreen}{#1}%
  \else
    #1%
  \fi
}

\newcommand{\revisiondelete}[1]{%
  \ifrevisions
    \textcolor{red}{#1}%
  \else
  \fi
}

\newcommand{\revisionchange}[2]{%
  \ifrevisions
    \textcolor{red}{#1}%
    \textcolor{darkgreen}{#2}%
  \else
    #2%
  \fi
}

\begin{document}

\title{CaRLi-V: Camera-RADAR-LiDAR Point-Wise 3D Velocity Estimation\looseness-1}

\author{Landson~Guo$^*$, Andres~M.~Diaz~Aguilar$^*$, William~Talbot$^{\ddagger\dagger}$, Turcan~Tuna$^{\ddagger}$, Marco Hutter$^{\ddagger}$, Cesar Cadena$^{\ddagger}$
\thanks{$^*$Authors are with ETH Z\"urich, Z\"urich, Switzerland and contributed equally.}
\thanks{$^{\ddagger}$ Authors are with the Robotic Systems Lab (RSL), ETH Z\"urich.}
\thanks{$^{\dagger}$ Corresponding author (\textit{E-mail address:} wtalbot@ethz.ch).}
\thanks{The research leading to these results has received funding from armasuisse Science and Technology and Swiss National Science Foundation (SNSF) under grant CHIST-ERA-23-MultiGIS-07.}
}

\markboth{ IEEE Robotics and Automation Practice (RA-P)}%
{Guo \MakeLowercase{et al.}, CaRLi-V: Camera-RADAR-LiDAR Point-Wise 3D Velocity Estimation}

\maketitle

\begin{abstract}
Accurate point-wise velocity estimation in 3D is crucial for robot interaction with non-rigid\revisiondelete{,} dynamic agents, \revisiondelete{such as humans, }enabling robust performance in path planning, collision avoidance, and object manipulation in dynamic environments. To this end, this paper proposes a novel RADAR, LiDAR, and camera fusion pipeline for point-wise 3D velocity estimation named CaRLi-V. This pipeline leverages raw RADAR measurements to create a novel RADAR representation, the velocity cube, which densely \revisionchange{represents }{encodes RADAR} radial velocities\revisiondelete{ within the RADAR's field-of-view}. By combining the velocity cube for radial velocity extraction, optical flow for tangential velocity estimation, and LiDAR for point-wise range measurements through a closed-form solution, our approach can produce 3D velocity estimates for a dense array of points. Developed as an open-source ROS2 package\footnote{\url{https://github.com/Soldann/CaRLi-V}\label{footnote_github}}, CaRLi-V has been field-tested \revisionchange{against }{on} a custom dataset \revisiondelete{and proven to produce low velocity error metrics relative to ground truth, enabling point-wise velocity estimation for robotic applications }\revisionadd{and achieves low velocity error metrics relative to ground truth while outperforming state-of-the-art scene flow methods}.\looseness-1


\end{abstract}


\vspace{-0.4cm}
\section{Introduction}

3D velocity estimation is \revisiondelete{an } essential \revisionchange{component in enabling }{for} robotic applications in dynamic environments, enabling systems to perceive and react effectively to \revisionchange{dynamic }{moving} objects. Conventional 3D velocity estimation \revisiondelete{methods }\revisionchange{tend }{tends} to rely on object-wise estimates, which can be inaccurate in the case of non-rigid objects that are partially in motion \cite{RADARnet, bi-lrfusion, lirafusion, radar_velocity_estimation_from_object_detection, camera_vel_1, camera_vel_2}. In contrast, dense velocity estimates consist of per-point velocity measurements that describe the motion of all surfaces in a scene. This captures fine-grained motion variations that can improve downstream tasks such as object detection and collision avoidance in dynamic environments~\cite{RADAR_full_velocity}.
\revisionchange{Nevertheless, extracting these velocities in a dense representation can be a challenging task requiring specialized hardware. }{However, extracting dense velocities remains challenging and often requires specialized hardware.}
Sensors such as Frequency-Modulated Continuous-Wave (FMCW) RADAR and \revisiondelete{FMCW }LiDAR enable the direct extraction of radial velocity. However, RADAR data is difficult to interpret directly and presents signal processing challenges, such as multi-path reflections, low signal-to-noise ratios, and limited resolving power, factors that explain why common detection algorithms, like Constant False Alarm Rate (CFAR), typically yield sparse point clouds~\cite{rohling2007radar}. While FMCW LiDAR can resolve many of these signal processing problems, it is a costly emerging technology that is not readily available~\cite{zhou2022towards,brune2024survey} while still limited to radial velocities. Alternatively, optical flow and scene flow algorithms extract velocity estimates from consecutive motion in indirect sensors such as cameras and Time of Flight (ToF) LiDAR, but these are either limited to tangential motion or \revisionchange{very computationally demanding }{incur very high computational cost}~\cite{battrawy2019dense,ng2021uncertainty}.

\begin{figure}[t]
    \centering    
    \includegraphics[width=\linewidth]{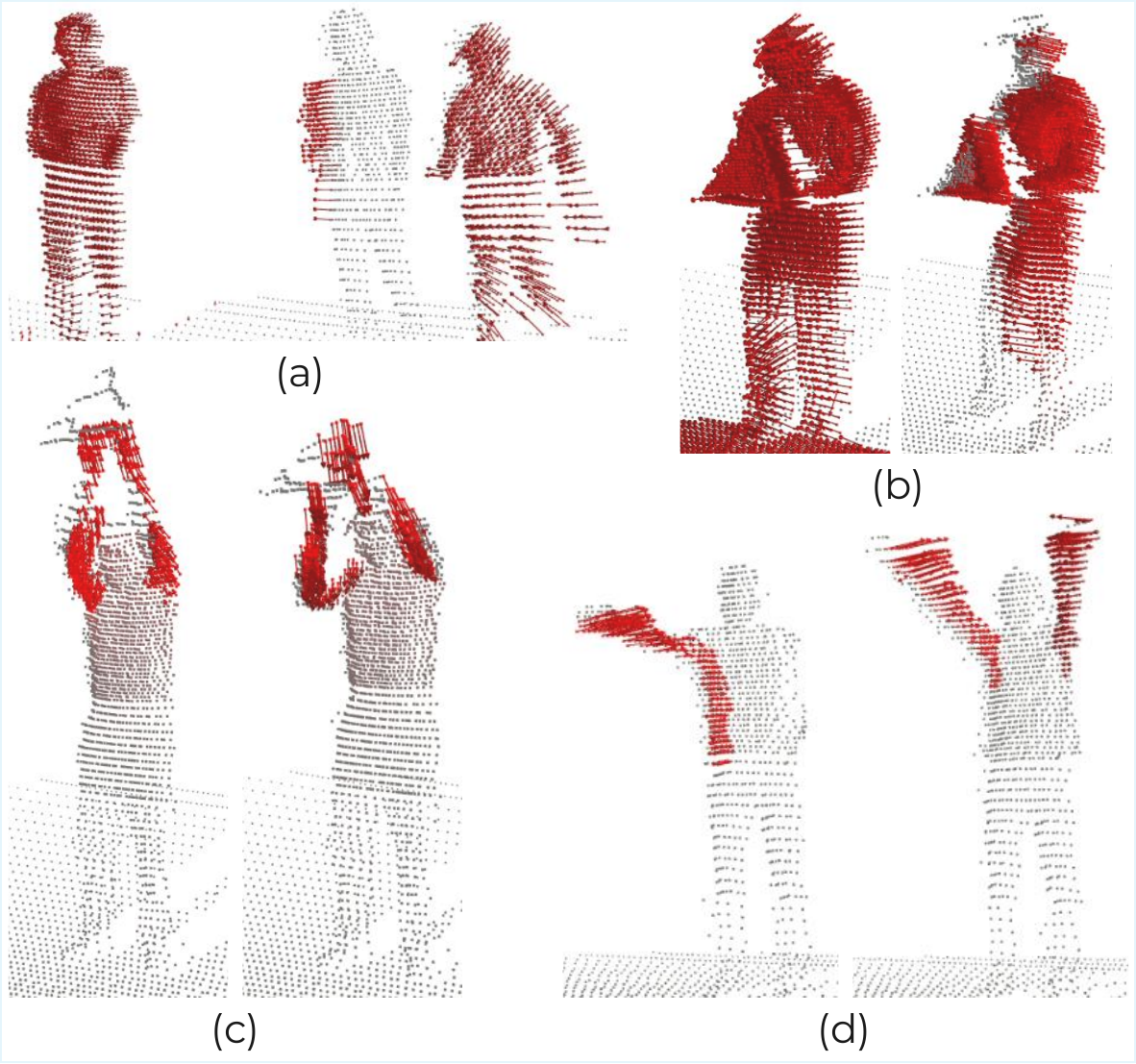}
    \vspace{-0.8cm}
    \caption{Resulting point clouds augmented with full velocity vectors displayed in red. The pipeline is able to (a) discern velocities between different dynamic agents; (b) estimate both radial and tangential velocities, as well as combinations of both, and (c, d) extract velocities from single parts of non-rigid moving agents, such as individual limbs in humans.}
    \label{fig: results}
    \vspace{-0.5cm}
\end{figure}

In this work, we propose a point-wise velocity estimation approach using perceptual sensor fusion from camera, RADAR, and LiDAR sensors. We build upon classical RADAR processing theory\revisiondelete{,} based on the RADAR cube\revisiondelete{,} to develop a novel representation, the velocity cube, that densely models radial velocities in space. Using the high spatial accuracy of the LiDAR points as a reference, we compute the velocity of each point by combining radial measurements from the velocity cube and tangential optical flow readings coinciding with these points to assign full velocity estimates. In this manner, CaRLi-V (Camera-RADAR-LiDAR point-wise Velocity estimation) is capable of estimating point-wise 3D velocities for all LiDAR points within camera and RADAR range as \revisionchange{illustrated }{shown} in Fig~\ref{fig: results}, following the recent trend of cross-modal motion estimation using Doppler cues~\cite{khoche2025dogflow}. We implement this approach as an open-source ROS package\textsuperscript{\ref{footnote_github}} and validate its performance in real-world settings using a custom dataset that has been provided to facilitate the reproduction of our results.\looseness-1

\vspace{-0.3cm}
\section{Related Work}
\label{sec:rel-work}

Recent works on full velocity estimation from direct measurements predominantly focus on object-wise estimates~\cite{RADARnet, bi-lrfusion, lirafusion, radar_velocity_estimation_from_object_detection, camera_vel_1, camera_vel_2}, restricting predictions to detected rigid objects. To tackle these limitations, Long et al.~\cite{RADAR_full_velocity} propose a method for point-wise full velocity estimation by augmenting RADAR point cloud measurements with camera-based optical flow for tangential velocity estimation. They present a closed-form geometric solution for the full velocity, as well as a neural network to correct for RADAR positional inaccuracies. However, relying on the RADAR point cloud introduces sparsity in the resulting velocity estimates. 
Scene flow directly leverages LiDAR's higher spatial resolution to produce denser velocity estimates by computing displacement vectors between consecutive point cloud measurements, with efficient implementations including \revisionchange{ICP Flow~\cite{ICP-FLow}, SeFlow~\cite{SeFlow}, and Fast Neural Scene Flow~\cite{FastNeuralSceneFlow}}{\cite{ICP-FLow, SeFlow, FastNeuralSceneFlow, deltaflow, flow4d, teflow}}. 
However, in addition to being computationally expensive, since scene flow is estimated instead of directly observed, it has unbounded error that potentially exceeds RADAR measurement noise in complex scenes.\looseness-1

As demonstrated by Long et al.~\cite{RADAR_full_velocity}, optical flow, used to estimate pixel-wise displacement vectors, can be leveraged to estimate tangential velocities. Many dense optical flow algorithms are inspired by RAFT~\cite{RAFT}, with notable extensions \revisionchange{such as SEA-RAFT~\cite{SEA-RAFT}, MS-RAFT+~\cite{MS-RAFT}, and RAFT-3D~\cite{RAFT3D} }{including \cite{SEA-RAFT, MS-RAFT, RAFT3D}}. While accurate, these algorithms \revisionchange{suffer from poor computation times due to their complexity }{are very computationally demanding}. To address this, alternatives such as NeuFlow~\cite{NeuFlow_v1,NeuFlow_v2} employ simpler architectures, enabling real-time performance with minimal loss of accuracy. \revisiondelete{Sparse optical flow algorithms, which estimate motion only for selected keypoints, present equally notable speed improvements. These often employ fast feature matchers\revisiondelete{such as XFeat}~\cite{XFeat} or more traditional methods such as KLT~\cite{KLT}.}

RADAR point clouds are the result of detection algorithms that focus on identifying strong intensity peaks over noise. While effective for robust target detection, this process discards weaker returns and thus loses substantial information, leading to sparse radial velocity measurements~\cite{rohling2007radar, richards2010principles}. A promising alternative is the RADAR cube, a 3D/4D object representing range, Doppler velocity, azimuth (and elevation) angle, computed using the Fast Fourier Transform (FFT) across all Analog-to-Digital Converter (ADC) dimensions, yielding a dense representation of RADAR returns. \revisionchange{Despite its promise, few studies directly exploit the RADAR cube. Most use it for object detection by cropping regions around points of interest~\cite{object_detection_radar_cube, RadarLidarDeepFusionDopplerContexts} or by projecting the cube into camera and ground planes for fusion with camera measurements~\cite{DPFT}. These works show that the RADAR cube encodes local motion patterns informative for dense velocity estimation and object detection }{Existing studies use the RADAR cube for object detection~\cite{object_detection_radar_cube, RadarLidarDeepFusionDopplerContexts, DPFT} and demonstrate its capabilities for encoding local motion patterns, but none to the authors' knowledge extend upon its capacities for dense velocity estimation}. \looseness-1

Recent advances in scene flow techniques focus on integrating different sensor modalities to further improve velocity estimates in challenging scenarios. DoGFlow~\cite{khoche2025dogflow} introduces cross-modal supervision via RADAR Doppler measurements \revisionchange{to generate pseudo-labels for LiDAR scene flow, reducing reliance on expensive ground truth and improving performance in long-range and adverse weather scenarios. }{for LiDAR scene flow;} MilliFlow~\cite{milliFlow2023} focuses on estimating scene flow directly from RADAR point clouds\revisiondelete{, and demonstrates how Doppler information and learning can partially overcome radar sparsity and noise constraints}\revisionchange{. Additionally, }{; and} CamLiFlow~\cite{camliflow2023} integrates dense image and sparse LiDAR modalities in an end-to-end fusion framework\revisionchange{; their bidirectional fusion connections help handle non-rigid motion and improve accuracy with fewer parameters. These developments highlight emerging paradigms: (1) using Doppler or RADAR velocity cues not only for radial velocity but as supervisory signals or pseudo-labels for richer motion fields; (2) scene flow methods that fuse multiple modalities (RADAR, LiDAR, camera) to mitigate each sensor’s weaknesses; (3) focus on long-range, adverse weather, and non-rigid motion scenarios, which are especially challenging. }{. These demonstrate how camera and Doppler information can help with non-rigid motion handling, improving accuracy and highlighting the effectiveness of cross-modal fusion in addressing the limitations of LiDAR-only scene flow.}\looseness-1

\begin{figure*}[t]
    \centering    
    \includegraphics[width=\linewidth]{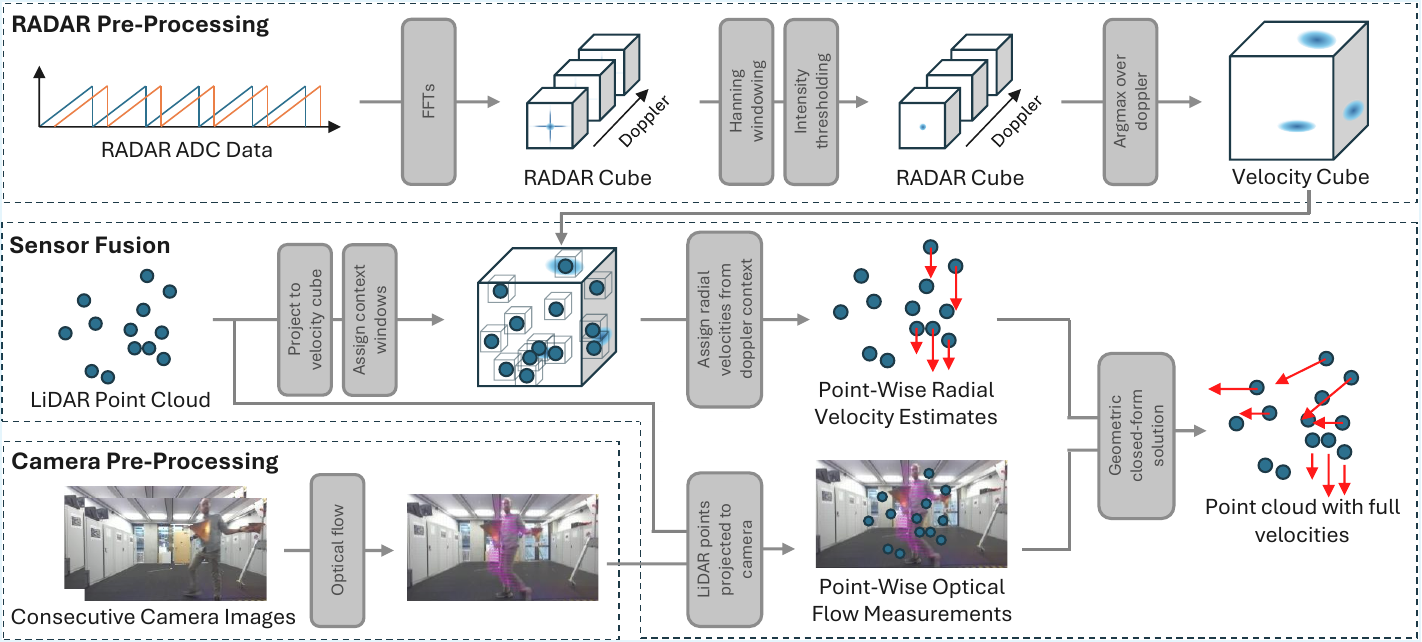}
    \vspace{-0.6cm}
    \caption{The CaRLi-V pipeline is divided into three steps: \emph{RADAR preprocessing}, where raw ADC RADAR data is used to compute the RADAR velocity cube, a dense representation of velocities in the environment; \emph{Camera preprocessing}, where two consecutive camera images are used to compute optical flow vectors for each pixel in the image; \emph{Sensor fusion}, where the LiDAR point cloud is projected into both representations to extract radial velocity and optical flow readings, with both estimates combined through a closed-form solution.}
    \label{fig:pipeline}
    \vspace{-0.6cm}
\end{figure*}

\vspace{-0.2cm}
\section{Methodology}
\label{sec:approach}

\subsection{RADAR preprocessing}

Traditionally, 4D RADAR processing involves applying \revisiondelete{a sequence of }FFTs along the sample, chirp, and antenna dimensions of the raw ADC data to construct a 4D RADAR cube, a dense representation of RADAR returns \revisionchange{where each axis corresponds to }{with axes corresponding to} range, Doppler, azimuth, and elevation. However, due to significant noise present in raw RADAR measurements, \revisionchange{CFAR is commonly employed to suppress noise and isolate strong RADAR reflections~\cite{rohling2007radar}. CFAR identifies local peaks within the cube, enabling the extraction of discrete point returns used to generate RADAR point clouds~\cite{radar_review}. }{CFAR is commonly used to isolate strong reflections in the cube and extract discrete point returns for RADAR point cloud generation~\cite{rohling2007radar, radar_review}.\looseness-1}

However, peak-detection approaches discard useful information about object shape, size, and spatial continuity contained in the full RADAR cube. To preserve this information, we introduce a novel RADAR representation—the velocity cube—which encodes dense radial velocity across the measurement space. The velocity cube is a discretized 3D tensor with angular dimensions (range, azimuth, and elevation), where each voxel is assigned a single radial velocity value by collapsing the Doppler axis.
The velocity cube is computed for each spatial coordinate \((r, \alpha, \theta)\) as the radial velocity corresponding to the Doppler bin with maximum magnitude:
\begin{equation}\label{eq:velocity_cube}
    V(r, \alpha, \theta) = v_{d^*}, \quad 
    d^* = \underset{d}{\arg\max}\; R(r, \alpha, \theta, d),
\end{equation}
where \(R(r, \alpha, \theta, d)\) denotes the magnitude of the RADAR return at range \(r\), azimuth \(\alpha\), elevation \(\theta\), and Doppler index \(d\), and \(v_{d^*}\) is the radial velocity associated with bin \(d^*\).

To suppress noise, windowing is applied during FFT processing to mitigate spectral leakage caused by signal truncation during sampling, which breaks the assumption of signal periodicity in finite-length sequences \cite{windowing}. \revisiondelete{This introduces artificial frequency components that redistribute the signal response across a larger frequency range. }In the RADAR cube, this appears as stretched reflections along the \revisionchange{Doppler, range, and angular }{its} dimensions. A Hanning window is employed to taper the signal smoothly at both ends, reducing sidelobe levels and suppressing cross-shaped artifacts from strong point returns.

\begin{figure}[t]
    \centering
    \vspace{-0.2cm}
        \includegraphics[width=\linewidth]{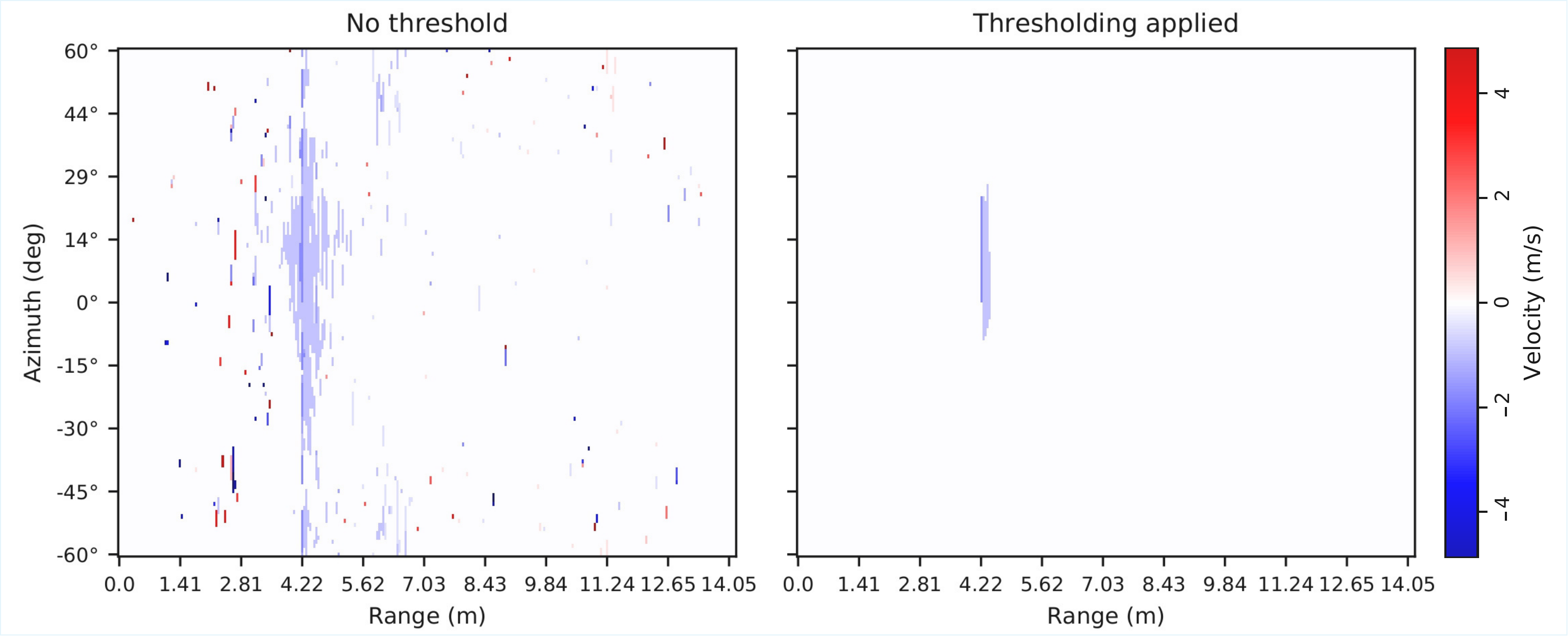}
        \vspace{-2.3em} 
        \caption{Effect of thresholding on the velocity cube. Thresholding removes salt-and-pepper noise and improves spatial precision along the angular dimensions, concentrating velocity readings to regions corresponding to moving agents.}
        \label{fig:noise_filtering}
        \vspace{-0.4cm}
\end{figure}

Furthermore, \revisionchange{relative intensity thresholding }{non-maximum suppression} is applied at the RADAR cube level to filter out values more than \revisionchange{\SI{5}{\decibel} }{\SI{16}{\decibel}} below the peak, enabling the removal of low-frequency noise while retaining spatial resolution. This reduces ambiguities in the location of objects along the angular dimensions, which is important given their lower resolution relative to range and Doppler in most RADAR. Furthermore, this removes the presence of salt-and-pepper noise stemming from low-intensity, low signal-to-noise ratio Doppler readings, which get carried into the velocity cube due to the argmax in Eq.~\ref{eq:velocity_cube}. After \revisionchange{thresholding }{suppression}, velocity readings remain concentrated only in regions corresponding to actual moving agents, as visible in Fig~\ref{fig:noise_filtering}. \revisionadd{The non-maximum suppression threshold follows from a parameter ablation study (see supplementary material), where \SI{16}{\decibel} achieved the lowest error by balancing noise suppression with the preservation of informative signal content.} \looseness-1
\vspace{-0.3cm}
\subsection{Camera Preprocessing}

In parallel \revisiondelete{to the RADAR preprocessing}, dense optical flow is \revisionchange{calculated on each }{computed between} consecutive \revisiondelete{camera }image \revisionchange{pair }{pairs} to obtain pixel-wise displacement vectors. For this, we make use of Neuflow v2~\cite{NeuFlow_v2} \revisionchange{due to }{for} its inference speed, but alternatives like RAFT~\cite{RAFT} could also be employed. \revisiondelete{These readings will be combined with depth data from LiDAR measurements to estimate the tangential velocity of each LiDAR point.}
\vspace{-0.6cm}
\subsection{Sensor fusion}
\revisionchange{After preprocessing the RADAR and camera data (see Fig~\ref{fig:pipeline} for the full pipeline), radial and tangential velocity estimates are computed for each LiDAR point by projecting it into the velocity cube and camera image. For radial velocity estimation, each LiDAR point is first converted to polar coordinates and assigned to the nearest bin in the discretized velocity cube. The radial velocity at each point is then defined as the highest magnitude velocity within a local context window centered around the projected point to mitigate the effects of spatial uncertainty inherent in RADAR measurements. In our implementation, this context window spans 10 azimuth bins, 10 elevation bins, and 20 range bins, corresponding to a physical coverage of about \SI{20}{\degree}$\times$\SI{50}{\degree}$\times$\SI{0.938}{\metre}. Note that the high range in elevation is due to our RADAR sensor having poor elevation resolution. A similar approach is taken in assigning optical flow readings, as each projected point is assigned the optical flow vector corresponding to the pixel it overlaps with. Here, no context windows are needed due to the high spatial resolution of the camera and LiDAR. }{
As shown in Fig~\ref{fig:pipeline}, the preprocessed RADAR and camera data are individually fused with the LiDAR point cloud to provide point-wise radial velocity measurements and image-plane motion constraints, which are jointly used to recover the full 3D velocities. To estimate radial velocities, since the LiDAR provides accurate 3D positions but no Doppler, while the RADAR provides Doppler readings with inaccurate angles, we use the LiDAR points as "queries" into the RADAR velocity cube. 
For each LiDAR point, its coordinates are converted to polar form and used to assign it to the nearest bin in the discretized velocity cube. The radial velocity of that point is then defined as the maximum-magnitude velocity within a local context window to mitigate the effects of spatial uncertainty inherent in RADAR measurements.
In our implementation, this context window spans 5 azimuth bins, 10 elevation bins, and 20 range bins, corresponding to a physical coverage of about \SI{10}{\degree}$\times$\SI{50}{\degree}$\times$\SI{0.938}{\metre}. Once again, these were selected based on our parameter ablation study. 
}

\revisionadd{For tangential velocity estimation, each LiDAR point is projected onto the image, and its corresponding optical flow vector provides an image-space motion constraint for that point. Through the camera intrinsics, this pixel displacement is mapped to motion in normalized image coordinates, while the LiDAR-derived depth provides the scale required to convert this motion into metric displacement. Due to the high spatial resolution of the camera and LiDAR, no context window is required. These image constraints are combined with the RADAR-derived radial velocity in a closed-form geometric solution of Eq.~\ref{eq: closedformsolution} to recover the full 3D velocity vector. Originally derived by Long et al.~\cite{RADAR_full_velocity},} \revisiondelete{To combine the optical flow readings and the radial velocity estimates, we adopt the closed-form solution derived by Long et al.~\cite{RADAR_full_velocity} used to compute the velocity vector at each LiDAR point. Specifically,} this models the motion of a point from position $\boldsymbol{p}$ to $\boldsymbol{q}$ at constant velocity $\boldsymbol{\dot{m}}$ over time $\Delta t$. In this same interval, the camera moves from $B$ to $A$. The closed-form solution is \looseness-1
\begin{equation} \label{eq: closedformsolution}
^{A}\dot{\boldsymbol{m}} =
\left[
\begin{array}{c}
^{B}_{A}\boldsymbol{R}_{1} - u_p \, ^{B}_{A}\boldsymbol{R}_{3} \\
^{B}_{A}\boldsymbol{R}_{2} - v_p \, ^{B}_{A}\boldsymbol{R}_{3} \\
^{A}\boldsymbol{\hat{r}}^{\top}
\end{array}
\right]^{-1}
\left[
\begin{array}{c}
\left(^{B}\boldsymbol{q}_{1} - u_p \, ^{B}\boldsymbol{q}_{3} \right) / \Delta t \\
\left(^{B}\boldsymbol{q}_{2} - v_p \, ^{B}\boldsymbol{q}_{3} \right) / \Delta t \\
\dot{r}
\end{array}
\right],
\end{equation}

where $^{A}\dot{\boldsymbol{m}}$ is the 3D velocity of the point $\boldsymbol{q}$ in $A$ frame, $^{B}_{A}\boldsymbol{R}_{1}$, $^{B}_{A}\boldsymbol{R}_{2}$, $^{B}_{A}\boldsymbol{R}_{3}$ are the first, second, and third rows of the rotation matrix from $A$ to $B$, $^{B}\boldsymbol{q}_{1}$, $^{B}\boldsymbol{q}_{2}$,  $^{B}\boldsymbol{q}_{3}$ the coordinates of the point we are trying to measure in $B$ frame, $u_p$ and $v_p$ the image coordinates of point $\textbf{p}$, $^{A}\boldsymbol{\hat{r}}^{\top}$ is the unit vector pointing in the direction from the RADAR frame origin to point $q$ expressed in $A$ frame, and $\dot{r}$ the radial velocity measured by the RADAR. \revisionadd{For a detailed geometric derivation of how image-space motion is combined with radial velocity to recover full 3D velocity, the reader is referred to~\cite{RADAR_full_velocity}.}\looseness-1

\begin{figure*}
    \centering
    \includegraphics[width=\linewidth]{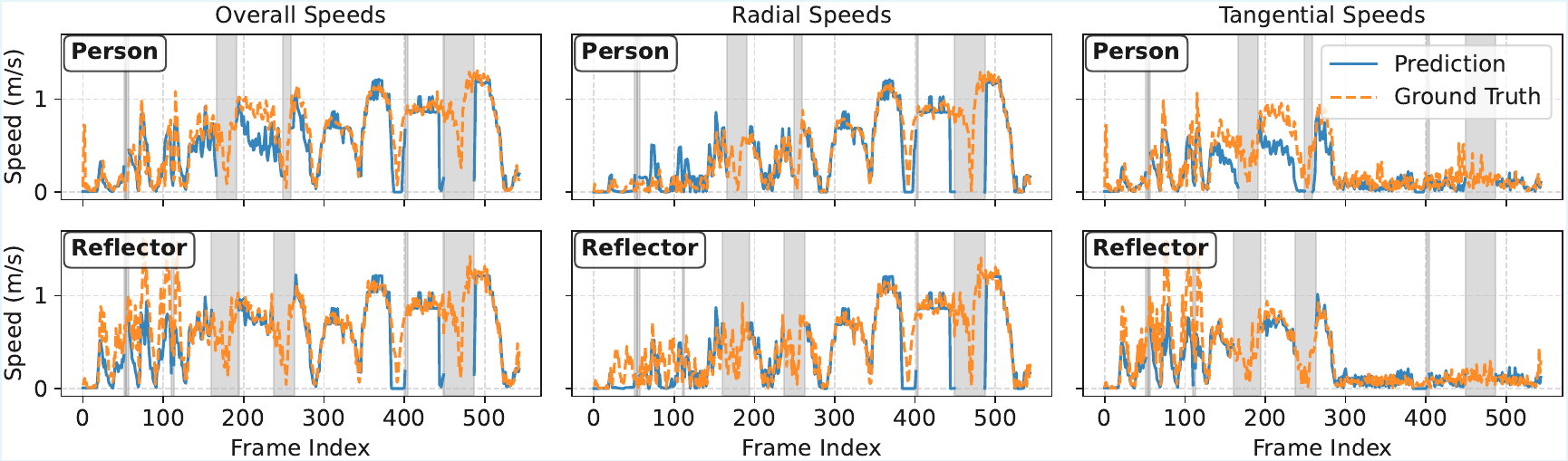}
    \vspace{-0.6cm}
    \caption{Plots of the magnitude of the estimated \revisionadd{(blue)} and ground truth \revisionadd{(orange)} velocity vectors, as well as their decompositions into radial and tangential components. These are taken from Scene 1 of the provided dataset. Gray regions show areas where the target object moved out sensor range. Note the jitter in the ground truth is due to the fact that we use instantaneous velocity estimates of the centroid. \looseness-1 }
    \label{fig:speed_plots}
    \vspace{-0.5cm}
\end{figure*}
\vspace{-0.2cm}
\section{Results \& Discussion}
\label{sec:exp}
Existing datasets providing RADAR, LiDAR, and camera data \revisiondelete{tend to }lack ground truth velocities~\cite{RobotCarDataset_1, RobotCarDataset_2, view_of_delft, astyx_dataset, nuscenes}, and those that do only provide RADAR \revisionchange{data in point cloud form }{point clouds}~\cite{nuscenes}. As such, we collected a new dataset consisting of \revisiondelete{various }scenes with humans walking, waving, and \revisionchange{moving RADAR retro-reflectors around }{manipulating RADAR retro-reflectors }to test our pipeline \revisionchange{with }{on} dynamic agents \revisionchange{performing }{undergoing} non-rigid \revisionchange{motion patterns }{motion}. \revisionchange{Access and further details about this dataset }{Details and access} are provided in the supplementary material.\looseness-1

Ground truth \revisionchange{data was derived from the collected dataset }{velocities were obtained} by \revisionchange{clustering objects and measuring their centroid displacement }{manually labelling bounding boxes around each object and measuring the displacement of their centroid} across consecutive frames. For direct comparison, the point-wise velocity estimates from our pipeline were averaged over the same object clusters. \revisionchange{Note that by averaging all point-wise velocities into a single estimate, the rigid object assumption is effectively reintroduced. }{
Note that this methodology incurs certain limitations: non-rigid motion (e.g., limb movement) can alter bounding box dimensions, shifting centroids and affecting the ground truth estimate, which explains the larger deviation observed for humans compared to the rigid reflector (see Fig~\ref{fig:speed_plots}). 
Additionally, the average of point-wise velocities may be biased by local motions that deviate from the overall object motion.} \revisionchange{However, point-wise to object-wise velocity conversion is straightforward for comparison purposes, and tracking the dominant motion of the entire object serves as a practical and robust proxy to evaluate the accuracy of our proposed pipeline. }{
Nevertheless, since centroid motion reflects the object translation, averaging point-wise velocities provides a coherent and comparable estimate by capturing the dominant motion shared across most points. While non-rigid motions introduce local deviations, these are typically spatially limited and directionally inconsistent, and therefore tend to cancel out or diminish under averaging.
}

\revisionchange{To quantify the pipeline's performance, the Average Velocity Error (AVE) was used, given by:\looseness-1
\begin{equation}
    \mathrm{AVE} = \frac{1}{N} \sum_{i=0}^N \left\| \mathbf{v}_i - \hat{\mathbf{v}}_i \right\|_2,
\end{equation}
where $N$ is the number of frames and $\mathbf{v}_i$, $\hat{\mathbf{v}}_i$ the ground truth and estimated object velocities at frame $i$ respectively. The AVE was also evaluated for tangential (AVE$_{\mathrm{tan}}$) and radial (AVE$_{\mathrm{rad}}$) velocity components. Furthermore, the Average Velocity Angular Error (AVAE) has been computed to evaluate the angular error of the estimated velocity vectors, given by:\looseness-1
\begin{equation}
    \mathrm{AVAE} = \frac{1}{N} \sum_{i=0}^N 
    \cos^{-1}\!\left(\frac{\mathbf{v}_i \cdot \hat{\mathbf{v}}_i}{\left\|\mathbf{v}_i\right\|_2 \cdot \left\|\hat{\mathbf{v}}_i\right\|_2}\right).
    \label{eq:AVAE}
\end{equation}
Additionally, a weighted AVAE (AVAE$_{\mathrm{w}}$) is computed by weighting the average in Eq.~\ref{eq:AVAE} by the magnitude of the ground truth velocity vector to weigh contributions with large velocities more, as small velocity vectors may be ambiguous in direction. To complement the mean-based metrics, median values were also computed for both the velocity error (MVE), including its tangential and radial components, and for the velocity angular error (MVAE). The implementation used to calculate these metrics is provided in the ROS package repository. }
{To quantify performance, we define the per-frame velocity error (VE) and velocity angular error (VAE) as:\looseness-1
\begin{equation}
    e_i = \left\| \mathbf{v}_i - \hat{\mathbf{v}}_i \right\|_2, \quad
    \theta_i = \cos^{-1}\!\left(\frac{\mathbf{v}_i \cdot \hat{\mathbf{v}}_i}{\left\|\mathbf{v}_i\right\|_2 \cdot \left\|\hat{\mathbf{v}}_i\right\|_2}\right).
\end{equation}
where $\mathbf{v}_i$ and $\hat{\mathbf{v}}_i$ are the ground truth and estimated velocities at frame $i$ respectively. We report the average (AVE, AVAE) and median (MVE, MVAE) of these quantities across frames, including tangential and radial components for the velocity error. Additionally, a weighted AVAE (AVAE$_{\mathrm{w}}$) is computed by weighting each $\theta_i$ by the magnitude of the ground truth velocity as small velocities may be ambiguous in direction.}

\revisionchange{Overall, the pipeline produces dense, high-resolution velocity estimates, distinguishing the motion of different parts of dynamic agents (e.g. limbs), and resolving multiple moving agents simultaneously (illustrated in Fig~\ref{fig: results}).}{Overall, the pipeline produces dense, high-resolution velocity estimates that distinguish the motion of different dynamic agents, resolve multiple moving agents simultaneously, and capture localized non-rigid motions such as human limb movement (see Fig.~\ref{fig: results}).} \revisiondelete{The pipeline demonstrates accurate object-wise velocity estimates with an overall AVE of \SI{0.22}{\meter\per\second}. The difference in AVE between human (\SI{0.236}{\meter\per\second}) and reflector targets (\SI{0.233}{\meter\per\second}) is not significant, so only overall results are shown in Table~\ref{tab:results}.} Notably, \revisiondelete{the }AVE$_{\mathrm{tan}}$ is higher than \revisiondelete{the }AVE$_{\mathrm{rad}}$, \revisionchange{pointing to weaknesses in the optical flow portion of the pipeline }{pointing to the optical flow branch as the main source of inaccuracy}. This is visible in Fig~\ref{fig:speed_plots}, particularly in frames 200-250, where the predicted tangential speed for the person is underestimated\revisionchange{. This likely stems from the choice of Neuflow v2 \cite{NeuFlow_v2} over slower but more accurate algorithms such as RAFT \cite{RAFT}}{, likely due to the use of NeuFlow v2~\cite{NeuFlow_v2} instead of slower but more accurate methods such as RAFT~\cite{RAFT}}. \revisionchange{Still, the gap remains small between both directions (\SI{0.04}{\meter\per\second}), indicating the pipeline is able to reliably capture velocities in all directions.}{Still, the gap between radial and tangential errors remains small, indicating that the proposed fusion strategy is able to recover motion reliably across different directions.} 

\begin{table}[t]
\vspace{-0.2cm}
\centering
\footnotesize
\setlength{\tabcolsep}{4pt}
\caption{Velocity estimation performance compared to the ground truth from scene 1 (standard deviation in parentheses)}
\vspace{-0.2cm}
\begin{tabular}{@{}lcccc@{}}
\toprule
Metric & TeFlow\cite{teflow} & Flow4D\cite{flow4d} & DeltaFlow\cite{deltaflow} & Ours \\
\midrule





\multicolumn{5}{l}{\textit{Velocity Magnitude Errors}} \\
\midrule
AVE [\SI{}{\meter\per\second}] 
& 0.39 (0.24) & 0.32 (0.23) & 0.32 (0.23) & \textbf{0.22 (0.21)} \\

AVE$_{\mathrm{rad}}$ [\SI{}{\meter\per\second}] 
& 0.17 (0.14) & 0.14 (0.13) & 0.15 (0.13) & \textbf{0.12 (0.16)} \\

AVE$_{\mathrm{tan}}$ [\SI{}{\meter\per\second}] 
& 0.33 (0.23) & 0.26 (0.22) & 0.25 (0.22) & \textbf{0.16 (0.16)} \\


MVE [\SI{}{\meter\per\second}] 
& 0.37 & 0.30 & 0.29 & \textbf{0.14} \\

MVE$_{\mathrm{rad}}$ [\SI{}{\meter\per\second}] 
& 0.13 & 0.11 & 0.12 &  \textbf{0.07} \\

MVE$_{\mathrm{tan}}$ [\SI{}{\meter\per\second}] 
& 0.29 & 0.22 & 0.2 & \textbf{0.11} \\

\midrule

\multicolumn{5}{l}{\textit{Velocity Angular Errors}} \\
\midrule
AVAE [\SI{}{\degree}] 
& 38.1 (33.0) &  40.8 (38.2) & 40.9 (38.1) & \textbf{28.2 (33.0)} \\

AVAE$_{\mathrm{w}}$ [\SI{}{\degree}] 
& 30.4 (28.4) & 30.0 (29.8) & 29.7 (29.0) & \textbf{18.0 (14.9)} \\

MVAE [\SI{}{\degree}] 
& 24.0 &  22.8 & 22.7 & \textbf{13.1} \\

\bottomrule
\end{tabular}
\label{tab:results}
\vspace{-0.5cm}
\end{table}

\revisionadd{
Table~\ref{tab:results} also shows that the standard deviations are relatively large for several metrics, indicating that the pipeline performs well for most frames, but degrades in a smaller number of more challenging cases.  This is consistent with the consistently lower median values, suggesting that the typical error is substantially lower than the mean error, while a limited number of difficult frames disproportionately increase the average. These cases are likely associated with ambiguous motion conditions such as very small target velocities, where angular errors become unstable and optical-flow-based tangential estimation becomes more sensitive to noise. This is also reflected by the gap between $\mathrm{AVAE}$ and $\mathrm{AVAE}_{\mathrm{w}}$, indicating that a substantial portion of the angular error is associated with low-speed points.} \looseness-1


\revisionadd{Compared to~\cite{teflow, flow4d, deltaflow}, CaRLi-V achieves lower error on all reported metrics in Table~\ref{tab:results}.
Since the compared baselines are learned scene flow methods that infer motion indirectly from consecutive LiDAR frames, while CaRLi-V combines LiDAR geometry with direct RADAR velocities and image-based tangential constraints, this highlights the value of the proposed cross-modal fusion strategy for dense velocity estimation.\looseness-1}


\revisionchange{It should be noted}{Note} that the RADAR's limited angular resolution combined with the large \revisionadd{Doppler} context \revisiondelete{window size }leads to velocity estimates occasionally bleeding into nearby static objects.
\revisionchange{This is at the origin of the difference between AVAE and AVAE$_{w}$ (shown in Table~\ref{tab:results}), indicating that a significant portion of the AVAE comes from misrepresenting static objects affected by the velocity bleeding. This effect can be mitigated using Kalman filtering or similar outlier rejection techniques. }
{This effect further increases the difference between AVAE and AVAE$_{w}$ since static points can exhibit large apparent angular deviations.\looseness-1
}

\vspace{-0.2cm}
\section{Conclusion}
\label{sec:conclusion}

CaRLi-V presents a novel field-tested \revisionchange{sensor}{RADAR, LiDAR, and camera} fusion pipeline for dense point-wise 3D velocity estimation\revisiondelete{using RADAR, LiDAR, and camera data}. By introducing the velocity cube and combining it with optical flow and LiDAR in closed-form, CaRLi-V enables accurate velocity estimation for dynamic, non-rigid agents\revisionchange{. Evaluated on a custom dataset, the pipeline shows strong agreement with object-wise ground truth }{, outperforming state-of-the-art scene flow methods when evaluated on a custom dataset}. \revisiondelete{Despite minor artifacts, resulting from the limited RADAR angular resolution the pipeline remains robust, and performance can be further improved with temporal filtering strategies. } Future work will focus on more advanced Doppler context window algorithms \revisionadd{and temporal filtering strategies} to reduce velocity bleeding, leveraging sparse optical flow for improved computational efficiency, and evaluating point-wise accuracy using motion capture techniques.\looseness-1

{
\bibliographystyle{IEEEtran}
\bibliography{references} 
}

\newpage

\begin{table*}
\centering
\footnotesize
\setlength{\tabcolsep}{8pt} 
\caption{Supplementary Material}
\begin{tabular}
{@{}p{0.16\linewidth}p{0.63\linewidth}p{0.13\linewidth}@{}}
\toprule

Supplementary Material & Description & Link \\ \midrule

Dataset & The dataset was collected on the sensor rig shown in our public repository (\url{https://github.com/Soldann/CaRLi-V}) using the following sensors:
\begin{itemize}
    \item V-MD3 RADAR transceiver
    \item HESAI QT128 LiDAR
    \item ZED 2i Camera
\end{itemize}
The dataset consists of various scenes, each saved as a separate MCAP file.  
Furthermore, an annotated version of scene 1 is provided along with an evaluation script, which was used to test the accuracy of the velocity estimates. This annotated scene includes \revisionchange{annotations }{bounding boxes} for the person and the RADAR reflector, which were moving throughout the scene. \revisionadd{The bounding boxes were manually labeled for each frame separately using the online data annotation platform "Supervisely" (\url{https://supervisely.com}).} The dataset consists of the following scenes \revisionadd{(including the number of LiDAR frames in parentheses)}:
\begin{itemize}
    \item \textit{Scene 1}: A person holding a radar reflector moving it up, down, left, right, forward, backward \revisionadd{(545 frames)}
    \item \textit{Scene 2}: A person moving forward, back, side to side, and in a figure 8 without a reflector \revisionadd{(448 frames)}
    \item \textit{Scene 3}: A person moving arms while stationary \revisionadd{(391 frames)}
    \item \textit{Scene 4}: A person moving just an arm holding the radar reflector \revisionadd{(336 frames)}
    \item \textit{Scene 5}: One stationary person holding a reflector, one moving person holding a reflector, and one moving then stationary person without a reflector \revisionadd{(304 frames)}
    \item \textit{Angle Tests 1-6}: A supplementary set of six MCAP files consisting of a person holding the RADAR reflector in different locations for testing of angular accuracy
\end{itemize}

\revisionadd{\textbf{Ground truth:} To compute the ground truth velocities, a Python script (included in our open-source repository) was used to extract the centroid of the bounding boxes at each frame, from which the person and the reflector velocities were computed by taking the displacement of the object's centroid between consecutive frames divided by the time between frames. To compare our pipeline's point-wise velocity estimates with these ground truth object-wise velocities, the average velocity of all points within each bounding box was used to compute the object-wise velocity at each frame. The same was done for the scene flow methods.}
& \revisionchange{\url{https://drive.google.com/drive/folders/1aJDNRScPyRDfbrbxfj-WOtP90-EwgWLH?usp=drive_link}}{\url{https://drive.google.com/drive/folders/1UL3VH1ohd_WElaz9_1rwBXL3sD56v-sE?usp=drive_link}} \\\midrule

Open-source package & The proposed pipeline has been implemented as an open-source ROS2 package. The package can be installed following the instructions on the GitHub page. & \url{https://github.com/Soldann/CaRLi-V} \\ \midrule

\revisionadd{Ablation study} & \revisionadd{An ablation study was conducted to determine the best performing parameters regarding the non-maximum suppression threshold and the Doppler context cube dimensions. These were chosen to be \SI{16}{\decibel} and 5 azimuth $\times$ 10 elevation $\times$ 20 range bins respectively. All experiments are provided in the linked document.} & \revisionadd{\href{https://drive.google.com/file/d/1mmHAB13cUg7lGWQMyjycJdf3t4YLK5_b/view?usp=drive_link}{Ablation Study Link}} \\ \midrule

Azimuth angle estimation & An explanation of the linear approximation that was used for angle estimation has been provided in the attached supplementary material. & \revisionadd{\href{https://drive.google.com/file/d/1mPeh8J50JbAY3uWld1gpxPougZ2WQW_p/view?usp=drive_link}{Azimuth Angle Estimation Link}} \\ \midrule

Supplementary Videos & \textit{Video 1:} Main video giving an overview of the pipeline and of the methods used in this work. 

& \url{https://youtu.be/k9M2VB4EQu8}\\

& \textit{Video 2:} Video of the pipeline running on Scenes 1 and 3 of the collected dataset illustrating the pipeline in action. These scenes display a person walking around and moving a RADAR reflector in the air, and a person moving their arms around while standing in place.

& \url{https://youtu.be/MKxTlQ3L6q0}\\ \\

& \textit{Video 3:} Video showcasing a comparison of the ground truth and computed velocity vectors in Scene 1. These are shown as velocity plots over time for the person and the retro-reflector individually.

& \url{https://youtu.be/MMEd8KCGGbM}\\ \\

& \textit{Video 4:} Video visualizing the effects of different windowing techniques on range-doppler RADAR cube data (of which the Hanning window was selected). This is compared to the RFFT (range FFT) output of the V-DM3 RADAR, which has its own signal processing pipeline to reject noise.

& \url{https://youtu.be/mqhvvih8nNs}\\ \\

& \textit{Video 5:} Video visualizing the velocity cube alongside the LiDAR data of a scene to better illustrate the concept of the velocity cube.

& \url{https://youtu.be/bcnT-UELoDg}\\
\bottomrule

\end{tabular}
\label{tab:supplementary}
\end{table*}

\begin{table*}[b]
\centering
\caption{Sensor settings and specifications used in our dataset.}
\label{tab:sensor-specs}
\resizebox{\textwidth}{!}{%
\begin{tabular}{@{}lc|lc|lc@{}}
\toprule
\multicolumn{2}{c}{\textit{V-MD3 RADAR}} & 
\multicolumn{2}{c}{\textit{HESAI QT128 LiDAR}} & 
\multicolumn{2}{c}{\textit{ZED 2i Camera}} \\ \midrule
Setting used & 7 & Settings used & -- & Settings used & -- \\
Max range [\si{\metre}] & \SI{6}{\metre} & Max range [\si{\metre}] & \SI{50}{\metre} & Depth range [\si{\metre}] & \SI{20}{\metre} (\SI{2.1}{\milli\metre}) / \SI{35}{\metre} (\SI{4}{\milli\metre}) \\
Max speed [\si{\kilo\metre\per\hour}] & \SI{10}{\kilo\metre\per\hour} & Max speed [\si{\kilo\metre\per\hour}] & -- & Max speed [\si{\kilo\metre\per\hour}] & -- \\
Range samples & 128 & Range samples & -- & Range samples & -- \\
Speed samples & 32 & Speed samples & -- & Speed samples & -- \\
Angle setting & 3D & FoV & \ang{360} $\times$ \ang{105.2} & FoV & \ang{110} $\times$ \ang{70} $\times$ \ang{120} (\SI{2.1}{\milli\metre}) / \ang{72} $\times$ \ang{44} $\times$ \ang{81} (\SI{4}{\milli\metre}) \\
Frame rate [\si{\milli\second}] & \SI{130}{\milli\second} & Frame rate [\si{\milli\second}] & \SIlist{100;50}{\milli\second} (\SIlist{10;20}{\hertz}) & Frame rate [\si{\milli\second}] & \SI{10}{\milli\second} (\SI{100}{\hertz}) \\
Range res. [\si{\centi\metre}] & \SI{4.69}{\centi\metre} & Range res. [\si{\centi\metre}] & -- & Range res. [\si{\centi\metre}] & -- \\
Speed res. [\si{\kilo\metre\per\hour}] & \SI{0.63}{\kilo\metre\per\hour} & Speed res. [\si{\kilo\metre\per\hour}] & -- & Speed res. [\si{\kilo\metre\per\hour}] & -- \\
\bottomrule
\end{tabular}%
}
\end{table*}

\end{document}